\pdfoutput=1
\documentclass{article}
\usepackage{ijcai22}

\usepackage{soul}
\usepackage{url}
\usepackage[hidelinks]{hyperref}
\usepackage[utf8]{inputenc}
\usepackage[small]{caption}
\usepackage{graphicx}
\usepackage{amsmath}
\usepackage{amsthm}
\usepackage{booktabs}
\usepackage{url}

\usepackage{multirow}
\usepackage{algorithmic}
\usepackage[linesnumbered,lined,boxed]{algorithm2e}
\usepackage{amsfonts}
\usepackage{thmtools}

\theoremstyle{theorem}
\newtheorem{theorem}{Theorem}
\newtheorem{theorem2}{Theorem}
\newtheorem{theorem3}{Theorem}

\newtheorem{example}[theorem]{Example}
\newtheorem{definition}[theorem2]{Definition}

\AtBeginDocument{\providecommand\BibTeX{{\normalfont B\kern-0.5em{\scshape i\kern-0.25em b}\kern-0.8em\TeX}}}

\usepackage{arydshln}

\usepackage{color, colortbl}
\definecolor{Gray}{gray}{0.9}

\newcolumntype{g}{>{\columncolor{Gray}}c}

\usepackage{thm-restate}
\usepackage[mathic=true]{mathtools}
\usepackage{fixmath}

\usepackage{pifont}

\usepackage{enumitem}
\setlist[enumerate]{noitemsep, topsep=0.5\topsep}
\setlist[description]{noitemsep, topsep=0.5\topsep}
\setlist[itemize]{noitemsep, topsep=0.5\topsep}

\usepackage{setspace}
\usepackage[protrusion=true,expansion=false, activate={true,nocompatibility},final,kerning=true,spacing=true,stretch=0, shrink=80]{microtype}

\usepackage{tablefootnote}

\usepackage{times}

\newcommand{\citet}[1]{\citeauthor{#1}~\shortcite{#1}}

\title{The Shapley Value in Machine Learning}
\author{
Benedek Rozemberczki$^1$ \and Lauren Watson$^2$ \and Péter Bayer$^3$  \and Hao-Tsung Yang$^2$\and\\ Oliver Kiss$^4$ \and Sebastian Nilsson$^1$ \And Rik Sarkar$^2$
\affiliations
$^1$Research Data \& Analytics, Research \& Development IT, AstraZeneca\\
$^2$The University of Edinburgh, School of Informatics\\
$^3$ Toulouse School of Economics \& Institute for Advanced Study in Toulouse\\
$^4$Central European University, Department of Economics and Business\\
\emails
benedek.rozemberczki@astrazeneca.com
}

\begin{document}

\maketitle

\begin{abstract}
Over the last few years, the Shapley value, a solution concept from cooperative game theory, has found numerous applications in machine learning. In this paper, we first discuss fundamental concepts of cooperative game theory and axiomatic properties of the Shapley value. Then we give an overview of the most important applications of the Shapley value in machine learning: feature selection, explainability, multi-agent reinforcement learning, ensemble pruning, and data valuation. We examine the most crucial limitations of the Shapley value and point out directions for future research.\footnote{The survey is supported by a collection of related work under \url{https://github.com/AstraZeneca/awesome-shapley-value}.
}

\end{abstract}

\section{Introduction}

Measuring importance and the attribution of various gains is a central problem in many practical aspects of machine learning such as explainability \cite{lundberg2017unified}, feature selection \cite{cohen2007feature}, data valuation \cite{ghorbani2019data}, ensemble pruning \cite{rozemberczki2021shapley} and federated learning \cite{wang2020principled,fan2021improving}. For example, one might ask: What is the importance of a certain feature in the decisions of a machine learning model? How much is an individual data point worth? Which models are the most valuable in an ensemble? These questions have been addressed in different domains using specific approaches. Interestingly, there is also a general and unified approach to these questions as a  solution to 
a \textit{transferable utility} (TU) cooperative game. In contrast with other approaches, solution concepts of TU games are theoretically motivated with axiomatic properties. The best known solution of this type is the \textit{Shapley value} \cite{shapley1953value} characterized by several  desiderata that include fairness, symmetry, and efficiency \cite{chalkiadakis2011computational}.

\begin{figure}[h!]
\centering 
\includegraphics[scale=0.09]{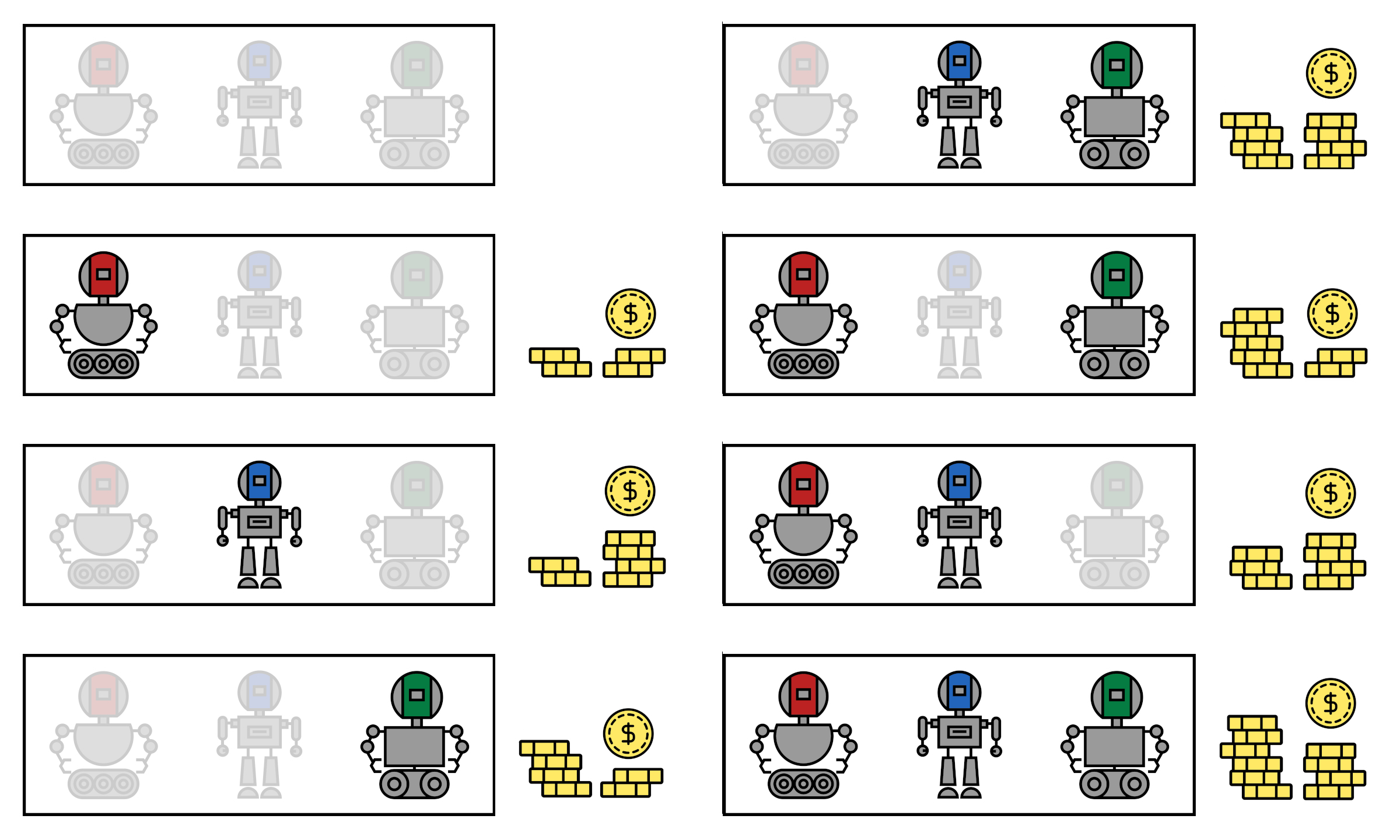}
\caption{The Shapley value can be used to solve cooperative games. An ensemble game is a machine learning application for it -- models in an ensemble are players (red, blue, and green robots) and the financial gain of the predictions is the payoff (coins) for each possible coalition (rectangles). The Shapley value can distribute the gain of the grand coalition (right bottom corner) among models. }\label{fig:eyecandy}
\end{figure}

In the TU setting, a cooperative game consists of: a \textit{player set} and a scalar-valued {\em characteristic function} that defines the value of \textit{coalitions} (subsets of players). In such a game, the Shapley value offers a rigorous and intuitive way to distribute the collective value (e.g. the revenue, profit, or cost) of the team across individuals.  To apply this idea to machine learning, we need to define two components: the player set and the characteristic function. In a machine learning setting {\em players} may be represented by a set of input features, reinforcement learning agents, data points, models in an ensemble, or data silos. The characteristic function can then describe the goodness of fit for a model, reward in reinforcement learning, financial gain on instance level predictions, or out-of-sample model performance. We provide an example about model valuation in an ensemble \cite{rozemberczki2021shapley} in Figure \ref{fig:eyecandy}.

\textbf{Present work.} We introduce basic definitions of cooperative games and present the Shapley value, a solution concept that can allocate gains in these games to individual players. We discuss its properties and emphasize why these are important in machine learning. We overview applications of the Shapley value in machine learning: feature selection, data valuation, explainability, reinforcement learning, and model valuation. Finally, we discuss the limitations of the Shapley value and point out future directions.

\section{Background}\label{sec:background}
This section introduces cooperative games and the Shapley value followed by its properties. We also provide an illustrative running example for our definitions.
\subsection{Cooperative games and the Shapley value}
\begin{definition}
\textbf{Player set and coalitions.} Let $\mathcal{N} = \{1, \dots , n\}$ be the finite set of players. We call each non-empty subset $\mathcal{S} \subseteq\mathcal{N}$ a \textit{coalition} and $\mathcal{N}$ itself the \textit{grand coalition}. 
\end{definition}
\begin{definition}
\textbf{Cooperative game.} A TU game is defined by the pair $(\mathcal{N}, v)$ where $v: 2^{\mathcal{N}} \rightarrow \mathbb{R}$ is a mapping called the \textit{characteristic function} or the \textit{coalition function} of the game assigning a real number to each coalition and satisfying $v(\emptyset)=0$.
\end{definition}
\begin{example}\label{example:one}
Let us consider a 3-player cooperative game where $\mathcal{N}=\left\{1,2,3 \right\}$.
The characteristic function defines the payoff for each coalition. Let these payoffs be given as:
{\scriptsize
\begin{alignat*}{4}
v(\emptyset) &= 0;& \,\,\,\, v(\left\{1\right\}) &= 7;&\,\,\,\, v(\left\{2\right\}) &= 11;&\,\,\,\, v(\left\{3\right\}) &= 14;\\  
v(\left\{1,2\right\}) &=18; & \,\,\,\, v(\left\{1,3\right\})  &=21;&\,\,\,\, v(\left\{2,3\right\}) &= 23;&\,\,\,\, v(\left\{1,2,3\right\}) &=25.
\end{alignat*}}
\end{example}
\begin{definition}\textbf{Set of feasible payoff vectors.} Let us define $\mathcal{Z}(\mathcal{N}, v)=\{\mathbf{z} \in \mathbb{R}^{\mathcal{N}} \mid \sum_{i \in \mathcal{N}} \textbf{z}_i \le v(\mathcal{N})\}$ the set of \textit{feasible payoff vectors} for the cooperative game $(\mathcal{N}, v)$.
\end{definition}
\begin{definition}
\textbf{Solution concept and solution vector.} \textit{Solution concept} $\Phi$ is a mapping associating a subset $\Phi(\mathcal{N}, v) \subseteq \mathcal{Z}(\mathcal{N}, v)$ to every TU game $(\mathcal{N}, v)$. A \textit{solution vector} $\phi(\mathcal{N}, v) \in \mathbb{R}^{\mathcal{N}}$ to the cooperative game $(\mathcal{N}, v)$ satisfies solution concept $\Phi$ if $\phi(\mathcal{N}, v) \in \Phi(\mathcal{N}, v)$. Solution concept $\Phi$ is \textit{single-valued} if for every $(\mathcal{N}, v)$ the set $\Phi(\mathcal{N},v)$ is a singleton.
\end{definition}
A solution concept defines an allocation principle through which rewards can be given to the individual players. The sum of these rewards cannot exceed the value of the grand coalition $v(\mathcal{N})$. Solution vectors are specific allocations satisfying the principles of the solution concept. 
\begin{definition}
\textbf{Permutations of the player set.} Let $\Pi(\mathcal{N})$ be the set of all permutations defined on $\mathcal{N}$, a specific permutation is written as $\pi \in \Pi(\mathcal{N})$ and $\pi(i)$ is the position of player $i\in \mathcal{N}$ in permutation $\pi$.
\end{definition}
\begin{definition} \textbf{Predecessor set.}
Let the set of predecessors of player $i\in \mathcal{N}$ in permutation $\pi$ be the coalition:
$$\mathcal{P}^{\pi}_i=\left\{j\in \mathcal{N}\mid \pi(j) < \pi(i) \right\}.$$
\end{definition}
Let us imagine that the permutation of the players in our illustrative game is $\pi=(3,2,1)$. Under this permutation the predecessor set of the $1^{st}$ player is $\mathcal{P}^{\pi}_1=\left\{3, 2\right\}$, that of the $2^{nd}$ player is $\mathcal{P}^{\pi}_2=\left\{3\right\}$ and $\mathcal{P}^{\pi}_3=\emptyset$.
\begin{definition}
\textbf{Shapley value.}\label{def:shapley} The Shapley value \cite{shapley1953value} is a single-valued solution concept for cooperative games. The $i^{th}$ component of the single solution vector satisfying this solution concept for any cooperative game $(\mathcal{N}, v)$ is given by Equation \ref{eq:shapley}.
{\small
\begin{align}
    \phi^{Sh}_i=\frac{1}{|\Pi (\mathcal{N})|}\sum\limits_{\pi \in \Pi(\mathcal{N})}\underbrace{[v(\mathcal{P}^{\pi}_i \cup \left\{ i\right\})-v(\mathcal{P}^{\pi}_i)]}_{\substack{\textbf{Player $i$'s marginal contribution } \\ \textbf{in permutation $\pi$}}}\label{eq:shapley}
\end{align}}
\end{definition}
The Shapley value of a player is the average marginal contribution of the player to the value of the predecessor set over every possible permutation of the player set. Table \ref{tab:shapley} contains manual calculations of the players' marginal contributions to each permutation and their Shapley values in Example \ref{example:one}.
\begin{table}[h!]
\caption{The permutations of the player set, marginal contributions of the players in each permutation and the Shapley values.}\label{tab:shapley}
\centering
{\footnotesize
\begin{tabular}{cccc}
            & \multicolumn{3}{c}{\textbf{Marginal Contribution} }\\ \hline
\textbf{Permutation} &\textbf{ Player 1 }    & \textbf{Player 2 }    & \textbf{Player 3}   \\ \hline
(1, 2, 3)     &     7        &     11         &     7         \\
(1, 3, 2)     &     7        &     4         &      14       \\
(2, 1, 3)     &     7         &    11        &      7         \\
(2, 3, 1)     &     2         &    11       &       12       \\
(3, 1, 2)     &     7        &     4         &      14      \\
(3, 2, 1)     &     2         &    9        &       14      \\ \hline
 \textbf{Shapley value}&       $32/6$       &     $50/6$         &    $68/6$         \\ \hline
\end{tabular}}
\end{table}

\subsection{Properties of the Shapley value}
We define the solution concept properties that characterize the Shapley value and emphasize their relevance and meaning in a \textit{feature selection game}. In this game input features are players, coalitions are subsets of features and the payoff is a scalar valued goodness of fit for a machine learning model using these input features.
\begin{definition}
\textbf{Null player.} Player $i$ is called a null player if $v(\mathcal{S} \cup \{i\}) = v(\mathcal{S}) \, \, \forall \mathcal{S} \subseteq \mathcal{N} \setminus \{i\}$. A solution concept $\Phi$ satisfies the null player property if for every game $(\mathcal{N}, v)$, every $\phi(\mathcal{N}, v) \in  \Phi(\mathcal{N}, v)$, and every null player $i$ it holds that $\phi(\mathcal{N}, v)_i=0$.
\end{definition}
In the feature selection game a solution concept with the null player property assigns zero value to those features that never increase the goodness of fit when added to the feature set.
\begin{definition}
\textbf{Efficiency.} A solution concept  $\Phi$ is \textit{efficient} or \textit{Pareto optimal} if for every game $(\mathcal{N}, v)$ and every solution vector $\phi(\mathcal{N}, v) \in  \Phi(\mathcal{N}, v)$ it holds that $\sum\limits_{i \in \mathcal{N}} \phi(\mathcal{N}, v)_i = v(\mathcal{N})$.
\end{definition}
Consider the goodness of fit of the model trained by using the whole set of input features. The importance measures assigned to individual features by an efficient solution concept sum to this goodness of fit. This allows for quantifying the contribution of individual features to the whole performance of the trained model.
\begin{definition}
\textbf{Symmetry.} Two players $i$ and $j$ are \textit{symmetric} if $v(\mathcal{S} \cup \{i\}) = v(\mathcal{S} \cup \{j\}) \, \, \forall \mathcal{S} \subseteq \mathcal{N} \setminus \{i, j\}$. A solution concept  $\Phi$ satisfies \textit{symmetry} if for all  $(\mathcal{N}, v)$ for all $\phi(\mathcal{N}, v) \in  \Phi(\mathcal{N}, v)$ and all symmetric players $i, j \in \mathcal{N}$ it holds that $\phi(\mathcal{N}, v)_i=\phi(\mathcal{N}, v)_j$.
\end{definition}
The symmetry property implies that if two features have the same marginal contribution to the goodness of fit when added to any possible coalition then the importance of the two features is the same. This property is essentially a fair treatment of the input features and results in identical features receiving the same importance score.
\begin{definition}
\textbf{Linearity.} A single-valued solution concept $\Phi$ satisfies \textit{linearity} if for any two games $(\mathcal{N},v)$ and $(\mathcal{N},w)$, and for the solution vector of the TU game given by $(\mathcal{N},v+w)$ it holds that
$$\phi(\mathcal{N}, v+w)_i = \phi(\mathcal{N}, v)_i + \phi(\mathcal{N}, w)_i,\,\,\,\,\forall i \in \mathcal{N}.$$
\end{definition}
Let us imagine a binary classifier and two sets of data points -- on both of these datasets, we can define feature selection games with binary cross entropy-based payoffs. The Shapley values of input features in the feature selection game calculated on the pooled dataset would be the same as adding together the Shapley values calculated from the two datasets separately.

These four properties together characterize the Shapley value.
\begin{theorem3}[Shapley, 1953]
A single-valued solution concept satisfies the \textit{null player}, \textit{efficiency}, \textit{symmetry}, and \textit{linearity} properties if and only if it is the Shapley value.
\end{theorem3}

\section{Approximations of the Shapley Value}\label{sec:approximations}

Shapley value computation requires an exponential number of characteristic function evaluations, resulting in exponential time complexity. This is prohibitive in a machine learning context when each evaluation can correspond to training a machine learning model. For this reason, machine learning applications use a variety of Shapley value approximation methods we discuss in this section. In the following discussion $\widehat{\phi}^{Sh}_i$ denotes an approximated Shapley value for player $i\in\mathcal{N}.$

\subsection{Monte Carlo Permutation Sampling}  Monte Carlo permutation sampling for the general class of cooperative games was first proposed by \citet{castro2009polynomial} to approximate the Shapley value in linear time. 

	\begin{algorithm}[h!]
{\small		\DontPrintSemicolon
		\SetAlgoLined
		\KwData{$(\mathcal{N},v)$ - Cooperative TU game.\\
	\,\,\,\,\,	\,\,\,\,\,\,\,\,\,$k$ - Number of sampled permutations.}
		\KwResult{$\widehat{\phi}^{Sh}_i$ - Approximated Shapley value $\forall i \in \mathcal{N}$.}

$\widehat{\phi}^{Sh}_i\leftarrow 0,\,\,\,\forall i\in\mathcal{N}$\;

\For{$(1,\dots,k)$}{
 $\pi \leftarrow \text{Uniform Sample}(\Pi(\mathcal{N}))$\;

\For{$i\in \mathcal{N}$}{

$\mathcal{P}^{\pi}_i\leftarrow \left\{j\in \mathcal{N}\mid \pi(j) < \pi(i) \right\}$\;
$\widehat{\phi}^{Sh}_i \leftarrow \widehat{\phi}^{Sh}_i+\frac{v(\mathcal{P}^{\pi}_i \cup \left\{ i\right\})-v(\mathcal{P}^{\pi}_i)}{k}$\;
}
}
}
		\caption{Monte Carlo permutation sampling approximation of the Shapley value.}
\label{alg:montecarlo}
\end{algorithm}

As shown in Algorithm~\ref{alg:montecarlo}, the method performs a sampling-based approximation. At each iteration, a random element from the permutations of the player set is drawn. The marginal contributions of the players in the sampled permutation are scaled down by the number of samples (which is equivalent to taking an average) and added to the approximated Shapley values from the previous iteration. \citet{castro2009polynomial} provide asymptotic error bounds for this approximation algorithm via the central limit theorem when the variance of the marginal contributions is known. \citet{maleki2013bounding}  extended the analysis of this sampling approach by providing error bounds when either the variance or the range of the marginal contributions is known via Chebyshev’s and Hoeffding’s inequalities. Their bounds hold for a finite number of samples in contrast to the previous asymptotic bounds.

\subsubsection{Stratified Sampling for Variance Reduction} In addition to extending the analysis of Monte Carlo estimation, \citet{maleki2013bounding}  demonstrate how to improve the Shapley Value approximation when sampling can be \textit{stratified} by dividing the permutations of the player set into homogeneous, non-overlapping sub-populations. In particular, they show that if the set of permutations can be grouped into strata with similar marginal gains for players, then the approximation will be more precise. Following this,   \citet{CASTRO2017180} explored stratified sampling approaches using strata defined by the set of all marginal contributions when the player is in a specific position within the coalition.  \citet{burgess2021approximating} propose stratified sampling approaches designed to minimize the uncertainty of the estimate via a stratified empirical Bernstein bound.

\subsubsection{Other Variance Reduction Techniques} Following the stratified approaches of \citet{maleki2013bounding,CASTRO2017180,burgess2021approximating}, \citet{illes2019estimation} propose an alternative variance reduction technique for the sample mean. Instead of generating a random sequence of samples, they instead generate a sequence of ergodic but not independent samples, taking advantage of negative correlation to reduce the sample variance. \citet{mitchell2021sampling} show that other Monte Carlo variance reduction techniques can also be applied to this problem, such as antithetic sampling~\cite{lomeli2019antithetic,rubinstein2016simulation}. A simple form of antithetic sampling uses both a randomly sampled permutation and its reverse. Finally, \citet{touati2021bayesian} introduce a Bayesian Monte Carlo approach to Shapley value calculation, showing that Shapley value estimation can be improved by using Bayesian methods to approximate the Shapley value. 
\subsection{Multilinear Extension}

By inducing a probability distribution over the subsets $\mathcal{S}$ where $\mathcal{E}_i$ is a random subset that does not include player $i$ and each player is included in a subset with probability $q$, \citet{owen1972multilinear} demonstrated that the sum over subsets in Definition ~\ref{def:shapley} can also be represented as an integral $\int_{0}^{1} e_i(q) dq$ where $e_i(q)=\mathbb{E}[v(\mathcal{E}_i \cup i)-v(\mathcal{E}_i)]$. Sampling over $q$ therefore provides an approximation method -- the multilinear extension. For example,~\citet{mitchell2021sampling} uses the trapezoid rule to sample $q$ at fixed intervals while \citet{okhrati2021multilinear} proposes incorporating antithetic sampling as a variance reduction technique. 

\subsection{Linear Regression Approximation}
In their seminal work \citet{lundberg2017unified} apply Shapley values to feature importance and explainability (SHAP values),  demonstrating that Shapley values for TU games can be approximated by solving a weighted least squares optimization problem. Their main insight is the computation of Shapley values by approximately solving the following optimization problem:

{\small
\begin{align}\label{kernelshap}
 w_\mathcal{S}&=\frac{|\mathcal{N}|-1}{{|\mathcal{N}| \choose |\mathcal{S}|}|\mathcal{S}|(|\mathcal{N}|-|\mathcal{S}|)} \\
\underset{\widehat{\phi}^{Sh}_0, ...,\widehat{\phi}^{Sh}_n}{min}\sum_{\mathcal{S}\subseteq \mathcal{N}}& w_\mathcal{S}\left(\widehat{\phi}^{Sh}_0+\sum_{i\in \mathcal{S}}\widehat{\phi}^{Sh}_i-v(\mathcal{S})\right) \label{obj}\\
s.t.\hspace{2mm} \widehat{\phi}^{Sh}_0=&v(\emptyset), \hspace{5mm}\widehat{\phi}^{Sh}_0+\sum_{i\in \mathcal{N}} \widehat{\phi}^{Sh}_i=v(\mathcal{N}).
\end{align}}

The definition of weights in Equation~\eqref{kernelshap} and the objective function in Equation \eqref{obj} implies the evaluation of $v(\cdot)$ for $2^n$ coalitions. To address this~\citet{lundberg2017unified} propose approximating this problem subsampling the coalitions.  Note that $w_\mathcal{S}$ is higher when coalitions are large or small. \citet{covert2021improving} extend the study of this method, finding that while SHAP is a consistent estimator, it is not an unbiased estimator. By proposing and analyzing a variation of this method that is unbiased, they conclude that while there is a small bias incurred by SHAP it has a significantly lower variance than the corresponding unbiased estimator. \citet{covert2021improving} then propose a variance reduction method for SHAP, improving convergence speed by a magnitude through sampling coalitions in pairs with each selected alongside its complement. 

\section{Machine Learning and the Shapley Value}\label{sec:applications}

\begin{table*}[h!]
\centering
\caption{An application area, payoff definition, Shapley value approximation technique, and computation time (the player set is noted by $\mathcal{N}$) based comparison of research works. Specific applications of the Shapley value are grouped together and ordered chronologically.}\label{tab:main}
{\scriptsize
\begin{tabular}{ccccc}
\hline
 \textbf{Application}& \textbf{Reference} &  \textbf{Payoff} & \textbf{Approximation} &\textbf{Time}\\
 \hline
\multirow{6}{*}{\textbf{Feature Selection}} & \cite{cohen2007feature}   & Validation loss & Exact  &$\mathcal{O}(|\mathcal{N}|!)$ \\
&\cite{sun2012feature} & Mutual information & Exact & $O(|\mathcal{N}|!)$\\
 & \cite{williamson2020efficient} &  Validation loss & Monte Carlo sampling & $\mathcal{O}(|\mathcal{N}|)$\\
& \cite{tripathi2020interpretable} &  Training loss & Monte Carlo sampling &$O(|\mathcal{N}|)$ \\
&\cite{patel2021game}  & Validation loss & Monte Carlo sampling &$O(|\mathcal{N}|)$\\
 &\cite{guha2021cga} & Validation loss& Exact & $O(|\mathcal{N}|!)$\\

\hline
\multirow{6}{*}{\textbf{Data Valuation}} & \cite{jia2019towards} & Validation loss  & Restricted Monte Carlo sampling & $\mathcal{O}(\sqrt{|\mathcal{N}|}\log |\mathcal{N}|^2)$ \\
 & \cite{ghorbani2019data} &  Validation loss & Monte Carlo sampling & $\mathcal{O}(|\mathcal{N}|)$ \\
  & \cite{shim2021online}  & Validation loss& Exact & $\mathcal{O}(|\mathcal{N}| \log |\mathcal{N}|)$ \\
& \cite{deutch}  & Validation loss  & Restricted Monte Carlo sampling  & $\mathcal{O}(|\mathcal{N}|)$ \\
 &  \cite{kwon2021efficient}  & Validation loss  & Monte Carlo sampling & $\mathcal{O}(|\mathcal{N}|)$ \\
 & \cite{kwon2021beta}  &  Validation loss  & Monte Carlo sampling & $\mathcal{O}(|\mathcal{N}|)$\\
\hline
\multirow{1}{*}{\textbf{Federated Learning}} & \cite{liu2021gtg}   & Validation loss &  Monte Carlo sampling&$\mathcal{O}(|\mathcal{N}|)$\\
\hline
 \multirow{7}{*}{\textbf{Universal Explainability}}   & \cite{lundberg2017unified} & Attribution & Linear regression &$\mathcal{O}(|\mathcal{N}|)$\\
  & \cite{sundararajan2020shapley}   & Interaction attribution & Integrated gradients &$\mathcal{O}(|\mathcal{N}|^2)$\\
  & \cite{sundararajan2020many} &Interaction attribution & Integrated gradients  &$\mathcal{O}(|\mathcal{N}|^2)$\\
   & \cite{frye2020shapley}& Attribution & Linear regression &$\mathcal{O}(|\mathcal{N}|)$ \\
     & \cite{frye2020asymmetric}& Attribution & Linear regression & $\mathcal{O}(|\mathcal{N}|)$\\
  & \cite{subgraphx_icml21} & Attribution & Monte Carlo sampling &$O(|\mathcal{N}|)$ \\
  &\cite{covert2021improving}&  Attribution& Linear regression& $\mathcal{O}(|\mathcal{N}|)$\\
 \hline
 \multirow{4}{*}{\textbf{Explainability of Deep Learning}}& \cite{chen2018shapley} & Attribution & Restricted Monte Carlo sampling  & $\mathcal{O}(2^{|\mathcal{N}|})$ or $\mathcal{O}(|\mathcal{N}|)$ \\
 & \cite{ancona2019explaining} & Neuron attribution& Voting game &$\mathcal{O}(|\mathcal{N}|^2)$ \\
 & \cite{ghorbani2020neuron} &  Neuron attribution & Monte Carlo sampling & $\mathcal{O}(|\mathcal{N}|)$\\
 & \cite{zhang2021interpreting}  & Interaction Attribution & Linear regression & $\mathcal{O}(|\mathcal{N}|)$\\
\hline
 \multirow{4}{*}{\textbf{Explainability  of Graphical Models}} 
 & \cite{liu2020shapley}  &  Attribution&Exact  &$\mathcal{O}(|\mathcal{N}|!)$\\
  & \cite{heskes2020causal} & Causal Attribution  & Linear regression &$\mathcal{O}(|\mathcal{N}|)$ \\
 & \cite{wang2021shapley} & Causal Attribution  & Linear regression &$\mathcal{O}(|\mathcal{N}|)$ \\
 & \cite{singal21a} &  Causal Attribution  & Linear regression &$\mathcal{O}(|\mathcal{N}|)$ \\
\hline
  \multirow{2}{*}{\textbf{Explainability in Graph Machine Learning}}&\cite{subgraphx_icml21}  &  Edge level attribution & Monte Carlo sampling&$\mathcal{O}(|\mathcal{N}|)$ \\

 &\cite{graphshapvx}  & Edge level attribution &Linear regression & $\mathcal{O}(|\mathcal{N}|)$\\
  \hline
  \multirow{2}{*}{\textbf{Multi-agent Reinforcement Learning}}&\cite{wang2021shaq}  &   Global reward & Monte Carlo sampling  & $\mathcal{O}(|\mathcal{N}|)$\\
& \cite{shapcredit}& Global reward & Monte Carlo sampling & $\mathcal{O}(|\mathcal{N}|)$\\
\hline
\textbf{Model Valuation in Ensembles} &  \cite{rozemberczki2021shapley}& Predictive performance &Voting game& $\mathcal{O}(|\mathcal{N}|^2)$ \\
 \hline
\end{tabular}
}
\end{table*}
Our discussion about applications of the Shapley value in the machine learning domain focuses on the formulation of the cooperative games, definition of the player set and payoffs, Shapley value approximation technique used, and the time complexity of the approximation. We summarized the most important application areas with this information in Table \ref{tab:main} and grouped the relevant works by the problem solved.

\subsection{Feature Selection} 
The feature selection game treats input features of a machine learning model as players and model performance as the payoff  \cite{guyon2003introduction,fryer2021shapley}. The Shapley values of features quantify how much individual features contribute to the model's performance on a set of data points.
\begin{definition} \textbf{Feature selection game.} Let the player set be $\mathcal{N}=\left\{1, \dots , n\right\}$, for $\mathcal{S}\subseteq \mathcal{N}$ the train and test feature vector sets are  $\mathcal{X}_{\mathcal{S}}^{\text{Train}}=\left\{\textbf{x}^{\text{Train}}_i|i \in \mathcal{S}\right\}$ and $\mathcal{X}_{\mathcal{S}}^{\text{Test}}=\left\{\textbf{x}^{\text{Test}}_i|i \in \mathcal{S}\right\}$. Let $f_\mathcal{S}(\cdot)$ be a machine learning model trained using $\mathcal{X}^{\text{Train}}_\mathcal{S}$ as input, then the payoff is $v(\mathcal{S})=g(\textbf{y}, \widehat{\textbf{y}}_{\mathcal{S}})$ where $g(\cdot)$ is a goodness of fit function, $\textbf{y}$ and  $\widehat{\textbf{y}}_{\mathcal{S}}=f_{\mathcal{S}}(\mathcal{X}_{\mathcal{S}}^{\text{Test}})$ are the ground truth and predicted targets.

\end{definition} 
Shapley values, and close relatives such as the Banzhaf index \cite{banzhaf1964weighted}, have been studied as a measure of feature importance in various contexts~\cite{cohen2007feature,sun2012feature,williamson2020efficient,tripathi2020interpretable}. Using these importance estimates, features can be ranked and selected or removed accordingly. This approach has been applied to various tasks such as vocabulary selection in natural language processing~\cite{patel2021game} and feature selection in human action recognition~\cite{guha2021cga}.  

\subsection{Data valuation} 
In the data valuation game training set data points are players and the payoff is defined by the goodness of fit achieved by a model on the test data. Computing the Shapley value of players in a data valuation game measures how much data points contribute to the performance of the model.

\begin{definition} \textbf{Data valuation game.} 
Let the player set be $\mathcal{N} = \{(\textbf{x}_i
, y_i) \ | \  1 \leq i \leq n \}$ where $\textbf{x}_i$ is the input feature vector and $y_i$ is the target. Given the coalition $\mathcal{S}\subseteq \mathcal{N}$ let $f_{\mathcal{S}}(\cdot)$ be a machine learning model trained on  $\mathcal{S}$. Let us denote the test set feature vectors and targets as $\mathcal{X}$ and $\mathcal{Y}$, given $f_{\mathcal{S}}(\cdot)$ the set of predicted labels is defined as $\widehat{\mathcal{Y}}=\left\{f_\mathcal{S}(\textbf{x})| \textbf{x} \in \mathcal{X}\right\}$.  Then the payoff of a model trained on the data points $\mathcal{S}\subseteq \mathcal{N}$ is $v(\mathcal{S})=g(\mathcal{Y},\widehat{\mathcal{Y}})$ where $g(\cdot)$ is a goodness of fit metric.
\end{definition}

The Shapley value is not the only method for data valuation -- earlier works used function utilization~\cite{koh2017understanding,sharchilev2018finding}, leave-one-out testing \cite{cook1977detection} and core sets \cite{dasgupta2009sampling}. However, these methods fall short when there are fairness requirements from the data valuation technique  ~\cite{jia2019towards,ghorbani2019data,kwon2021beta}. Ghorbani proposed a framework of utilizing Shapley value in a data-sharing system~\cite{ghorbani2019data}; \citet{jia2019towards} advanced this work with more efficient algorithms to approximate the Shapley value for data valuation. The distributional Shapley value has been discussed by \citet{ghorbani2020distributional} who argued that keeping privacy is hard during Shapley value computation. Their method calculates the Shapley value over a distribution which solves problems such as lack of privacy. The computation time of this can be reduced as \citet{kwon2021efficient} point out with approximation methods optimized for specific machine learning models.

\subsection{Federated learning}
A federated learning scenario can be seen as a cooperative game by modeling the data owners as players who cooperate to train a high-quality machine learning model \cite{liu2021gtg}.

\begin{definition}\textbf{Federated learning game.}
In this game players are a set of labeled dataset owners $\mathcal{N}=\left\{(\mathcal{X}_i,\mathcal{Y}_i)| 1 \leq i\leq n\right\}$ where $\mathcal{X}_i$ and $\mathcal{Y}_i$ are the feature and label sets owned by the $i^{th}$ silo. Let $(\mathcal{X},\mathcal{Y})$ be a labeled test set,  $\mathcal{S}\subseteq \mathcal{N}$ a coalition of data silos, $f_\mathcal{S}(\cdot)$ a machine learning model trained on  $\mathcal{S}$, and $\widehat{\mathcal{Y}}_{\mathcal{S}}$ the labels predicted by $f_\mathcal{S}(\cdot)$ on $\mathcal{X}$. The payoff of $\mathcal{S}\subseteq \mathcal{N}$ is $v(\mathcal{S})=g(\mathcal{Y},\widehat{\mathcal{Y}}_{\mathcal{S}})$ where $g(\cdot)$ is a goodness of fit metric.
\end{definition}
The system described by \citet{liu2021gtg} uses Monte Carlo sampling to approximate the Shapley value of data coming from the data silos in linear time. Given the potentially overlapping nature of the datasets, the use of configuration games could be an interesting future direction \cite{albizuri2006configuration}.

\subsection{Explainable machine learning}
\label{subsec:explainable}
In explainable machine learning the Shapley value is used to measure the contributions of input features to the output of a machine learning model at the instance level. Given a specific data point, the goal is to decompose the model prediction and assign Shapley values to individual features of the instance. There are universal solutions to this challenge that are model agnostic and designs customized for deep learning \cite{chen2018shapley,ancona2019explaining}, classification trees \cite{lundberg2017unified}, and graphical models \cite{liu2020shapley,singal21a}.
\subsubsection{Universal explainability}
A cooperative game for universal explainability is completely model agnostic; the only requirement is that a scalar-valued output can be generated by the model such as the probability of a class label being assigned to an instance. 

\begin{definition} \textbf{Universal explainability game.} Let us denote the machine learning model of interest with $f(\cdot)$ and let the player set be the feature values of a single data instance: $\mathcal{N} = \left\{x_i| 1\leq i \leq n\right\}$. The payoff of a coalition $\mathcal{S}\subseteq \mathcal{N}$ in this game is the scalar valued  prediction $v(\mathcal{S})=\widehat{y}_{\mathcal{S}}=f(\mathcal{S})$ calculated from the subset of feature values.

\end{definition}

Calculating the Shapley value in a game like this offers a complete decomposition of the prediction because the \textit{efficiency} axiom holds. The Shapley values of feature values are explanatory attributions to the input features and missing input feature values are imputed with a reference value such as the mean computed from multiple instances \cite{lundberg2017unified,covert2021improving}.  The pioneering Shapley value-based universal explanation method SHAP \cite{lundberg2017unified} proposes a linear time approximation of the Shapley values which we discussed in Section \ref{sec:approximations}. This approximation has shortcomings and implicit assumptions about the features which are addressed by newer Shapley value-based explanation techniques. For example, in \cite{frye2020shapley} the input features are not necessarily independent, \cite{frye2020asymmetric} restricts the permutations based on known causal relationships, and in \cite{covert2021improving} the proposed technique improves the convergence guarantees of the approximation. Several methods generalize SHAP beyond feature values to give attributions to first-order feature interactions \cite{sundararajan2020many,sundararajan2020shapley}. However, this requires that the player set is redefined to include feature interaction values.

\subsubsection{Deep learning}
In neuron explainability games neurons are players and attributions to the neurons are payoffs. The primary goal of Shapley value-based explanations in deep learning is to solve these games and compute attributions to individual neurons and filters \cite{ghorbani2020neuron,ancona2019explaining}. 

\begin{definition} \textbf{Neuron explainability game.}
Let us consider $f_{\textsc{IN}}(\cdot)$ the encoder layer of a neural network and  \textbf{x} the input feature vector to the encoder. In the neuron explainability game the  player set is $\mathcal{N} = f_{\textsc{IN}}(\textbf{x})=\left\{h_1,\dots,h_n\right \}$  - each player corresponds to the output of a neuron in the final layer of the encoder. The payoff of coalition $\mathcal{S}\subseteq \mathcal{N}$ is defined as the predicted output $v(\mathcal{S})=\widehat{y}_{\mathcal{S}}=f_{\textsc{OUT}}(\mathcal{S})$ where $f_{\textsc{OUT}}(\cdot)$ is the head layer of the neural network. 
\end{definition}

In practical terms, the payoffs are the output of the neural network obtained by masking out certain neurons. Using the Shapley values obtained in these games the value of individual neurons can be quantified. At the same time, some deep learning specific Shapley value-based explanation techniques have designs and goals that are aligned with the games described in universal explainability. These methods exploit the structure of the input data  \cite{chen2018shapley} or the nature of feature interactions \cite{zhang2021interpreting} to provide efficient computations of attributions. 

\subsubsection{Graphical models}
Compared to universal explanations the graphical model-specific techniques restrict the admissible set of player set permutations considered in the attribution process. These restrictions are defined based on known causal relations and permutations are generated by various search strategies on the graph describing the probabilistic model \cite{heskes2020causal,liu2020shapley,singal21a}. Methods are differentiated from each other by how restrictions are defined and how permutations are restricted.
\subsubsection{Relational machine learning}
In the relational machine learning domain the Shapley value is used to create edge importance attributions of instance-level explanations \cite{graphshapvx,subgraphx_icml21}. Essentially the Shapley value in these games measures the average marginal change in the outcome variable as one adds a specific edge to the edge set in all of the possible edges set permutations. It is worth noting that the edge explanation and attribution techniques proposed could be generalized to provide node attributions.

\begin{definition} \textbf{Relational explainability game.}
Let us define a graph $\mathcal{G}=(\mathcal{V},\mathcal{N})$ where $\mathcal{V}$ and  $\mathcal{N}$ are the vertex and edge sets. Given the relational machine learning model $f(\cdot)$, node feature matrix $\textbf{X}$, node $u \in \mathcal{V}$, the payoff of coalition $\mathcal{S} \subseteq \mathcal{V}$ in the graph machine learning explanation game is defined as the node level prediction $v(\mathcal{S}) = \widehat{y}_{\mathcal{S},u} = f(\textbf{X}, \mathcal{V}, \mathcal{S}, u)$. 
\end{definition}
\subsection{Multi-agent reinforcement learning}
Global reward multi-agent reinforcement learning problems can be modeled as TU games \cite{wang2021shaq,shapcredit} by defining the player set as the set of agents and the payoff of coalitions as a global reward. The Shapley value allows an axiomatic decomposition of the global reward achieved by the agents in these games and the fair attribution of credit assignments to each of the participating agents.

\subsection{Model valuation in ensembles} 
The Shapley value can be used to assess the contributions of machine learning models to a composite model in {ensemble games}. In these games, players are models in an ensemble and payoffs are decided by whether prediction mad by the model are correct.

\begin{definition} \textbf{Ensemble game}.
Let us consider a single target - feature instance denoted by $(y, \textbf{x})$. The player set in ensemble games is defined by a set of machine learning models $\mathcal{N}=\left\{f_i(\cdot) | 1 \leq i \leq n\right\}$ that operate on the feature set. The predicted target output by the ensemble $\mathcal{S}\subseteq \mathcal{N}$ is defined as $\widehat{y}_{\mathcal{S}} = \tilde{f}( \mathcal{S}, \textbf{x})$ where $\tilde{f}(\cdot)$ is a prediction aggregation function. The payoff of $\mathcal{S}$ is $v(\mathcal{S})=g(y, \widehat{y}_\mathcal{S})$ where $g(\cdot)$ is a goodness of fit metric.
\end{definition}
The ensemble games described by \cite{rozemberczki2021shapley} are formulated as a special subclass of voting games. This allows the use of precise game-specific approximation \cite{fatima2008linear} techniques and because of this the Shapley value estimates are obtained in quadratic time and have a tight approximation error. The games themselves are model agnostic concerning the player set -- ensembles can be formed by heterogeneous types of machine learning models that operate on the same inputs.

\section{Discussion}\label{sec:discussion}
The Shapley value has a wide-reaching impact in machine learning, but it has limitations and certain extensions of the Shapley value could have important applications in machine learning.
\subsection{Limitations}
\subsubsection{Computation time}
Computing the Shapley value for each player naively in a TU game takes factorial time. In some machine learning application areas such as multi-agent reinforcement learning and federated learning where the number of players is small, this is not an issue.  However, in large scale data valuation \cite{kwon2021efficient,kwon2021beta}, explainability \cite{lundberg2017unified}, and feature selection \cite{patel2021game} settings the exact calculation of the Shapley value is not tractable. In Sections \ref{sec:approximations} and \ref{sec:applications} we discussed approximation techniques proposed to make Shapley value computation possible. In some cases, asymptotic properties of these Shapley value approximation techniques are not well understood -- see for example \cite{chen2018shapley}.
\subsubsection{Interpretability}
By definition, the Shapley values are the average marginal contributions of players to the payoff of the grand coalition computed from all permutations \cite{shapley1953value}. Theoretical interpretations like this one are not intuitive and not useful for non-game theory experts. This means that translating the meaning of Shapley values obtained in many application areas to actions is troublesome \cite{kumar2020problems}. For example in a data valuation scenario: is a data point with a twice as large Shapley value as another one twice as valuable? Answering a question like this requires a definition of the cooperative game that is interpretable.

\subsubsection{Axioms do not hold under approximations}
As we discussed most applications of the Shapley value in machine learning use approximations. The fact that under these approximations the desired axiomatic properties of the Shapley value do not hold is often overlooked \cite{sundararajan2020many}. This is problematic because most works argue for the use of Shapley value based on these axioms. In our view, this is the greatest unresolved issue in the applications of the Shapley value.

\subsection{Future Research Directions}
\subsubsection{Hierarchy of the coalition structure}
The Shapley value has a constrained version called Owen value \cite{owen1977values} in which only permutations satisfying conditions defined by a \textit{coalition structure} - a partition of the player set - are considered. The calculation of the Owen value is identical to that of the Shapley value, with the exception that only those permutations are taken into account where the players in any of the subsets of the \textit{coalition structure} follow each other. In several real-world data and feature valuation scenarios even more complex hierarchies of the coalition, the structure could be useful. Having a nested hierarchy imposes restrictions on the admissible permutations of the players and changes player valuation. Games with such nested hierarchies are called level structure games in game theory. \cite{winter89} presents the Winter value a solution concept to level structure games - such games are yet to receive attention in the machine learning literature.

\subsubsection{Overlapping coalition structure}
Traditionally, it is assumed that players in a coalition structure are allocated in disjoint partitions of the grand coalition. Allowing players to belong to overlapping coalitions in configuration games \cite{albizuri2006configuration} could have several applications in machine learning. For example in a data-sharing - feature selection scenario multiple data owners might have access to the same features - a feature can belong to overlapping coalitions.

\subsubsection{Solution concepts beyond the Shapley value}
The Shapley value is a specific solution concept of cooperative game theory with intuitive axiomatic properties (Section \ref{sec:background}). At the same time it has limitations with respect to computation constraints and interpretability (Sections \ref{sec:approximations} and \ref{sec:discussion}). Cooperative game theory offers other solution concepts such as the \textit{core}, \textit{nucleolus}, \textit{stable set}, and \textit{kernel} with their own axiomatizations. For example, the \textit{core} has been used for model explainability and feature selection  \cite{Yan_Procaccia_2021}. Research into the potential applications of these solution concepts is lacking.

\section{Conclusion}\label{sec:conclusions}

In this survey we discussed the Shapley value, examined its axiomatic characterizations and the most frequently used Shapley value approximation approaches. We defined and reviewed its uses in machine learning, highlighted issues with the Shapley value and potential new application and research areas in machine learning.

\section*{Acknowledgements}
This research was supported by REPHRAIN: The National Research Centre on Privacy, Harm Reduction and Adversarial Influence Online (UKRI grant: EP/V011189/1). The authors would like to thank Anton Tsitsulin for feedback throughout the preparation of this manuscript. 

\small

\bibliographystyle{named}
\bibliography{bibliography}

\end{document}